%% file: DPPs.tex
\documentclass{article}

\usepackage[final]{nips_2016}

\usepackage[utf8]{inputenc} 
\usepackage[T1]{fontenc}    
\usepackage[colorlinks]{hyperref}       
\usepackage{url}            
\usepackage{booktabs}       
\usepackage{amsfonts,amsmath,amssymb}       
\usepackage{nicefrac}       
\usepackage{microtype}      

\input{preamble}

\hypersetup{
  colorlinks=true,       
  linkcolor=MPG,          
  citecolor=dred,        
  filecolor=mgra,      
  urlcolor=ora,           
}

\renewcommand{\V}{\mathbb{V}}

\newlength{\figwidth}
\newlength{\figheight}
\usetikzlibrary{external}
\tikzset{external/force remake=false}
\tikzsetexternalprefix{fig/}

\title{Exact Sampling from Determinantal Point Processes}

\author{
  Philipp Hennig\\
  Max Planck Institute for Intelligent Systems\\
  T\"ubingen, Germany \\
  \texttt{ph@tue.mpg.de} \\
  \And
  Roman Garnett \\
  Washington University in St.\ Louis \\
  St.\ Louis, MO, USA \\
  \texttt{garnett@wustl.edu} \\
}

\begin{document}

\maketitle

\begin{abstract}
  Determinantal point processes (DPPs) are an important concept in random matrix theory and combinatorics. They have also recently attracted interest in the study of numerical methods for machine learning, as they offer an elegant ``missing link'' between independent Monte Carlo sampling and deterministic evaluation on regular grids, applicable to a general set of spaces. This is helpful whenever an algorithm \emph{explores} to reduce uncertainty, such as in active learning, Bayesian optimization, reinforcement learning, and marginalization in graphical models. To draw samples from a DPP in practice, existing literature focuses on approximate schemes of low cost, or comparably inefficient exact algorithms like rejection sampling. We point out that, for many settings of relevance to machine learning, it is also possible to draw \emph{exact} samples from DPPs on continuous domains. We start from an intuitive example on the real line, which is then generalized to multivariate real vector spaces. We also compare to previously studied approximations, showing that exact sampling, despite higher cost, can be preferable where precision is needed.
\end{abstract}

\section{Introduction}
\label{sec:introduction}

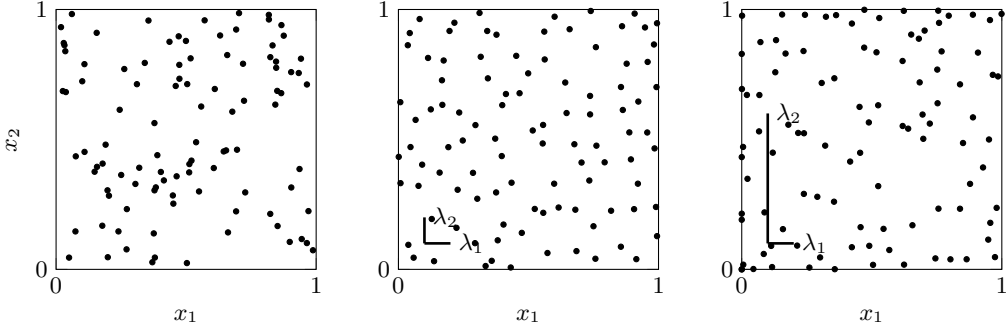
\begin{figure}[t]
  \centering\footnotesize
  \setlength{\figwidth}{0.9\textwidth}
  \input{fig/Samples}
  \caption{One hundred points sampled from, respectively, a uniform random (left) and two determinantal point processes (center, right)  with kernel $k_\text{SE}$~\eqref{eq:SE} of different length scales. Uniform random numbers exhibit ``clumping,'' randomly occurring patterns responsible for the stochastic convergence rate of Monte Carlo estimators. DPPs have a self-avoiding characteristic, yielding better coverage.}
\label{fig:samples}
\end{figure}

Determinantal point processes (DPPs), introduced by \citet{macchi1975coincidence}, are stochastic point processes whose joint probability measure is proportional to the determinant of a positive definite kernel Gram matrix (a more formal introduction follows in \textsection\ref{sub:determinantal_point_processes} below). Intuitively, this introduces a dependence between points sampled from such processes that gives them a ``repulsive'' property---point sets drawn from DPPs tend to cover a space more regularly than uniform random samples (cf.~Figure~\ref{fig:samples}).

DPPs initially arose in the study of fermionic gases in physics and have since found application in other areas, including random matrix theory \citep{mehta2004random}. A review of their statistical properties is provided by \citet{soshnikov2000determinantal}. DPPs have seen less attention in machine learning and statistics than in physics, but DPPs on discrete domains have been used as diversity-inducing priors \citep[e.g.,][]{kulesza2012determinantal}. An arguably under-explored direction is that they provide an elegant theoretical handle on the notion of \emph{exploration} that is of interest across machine learning. In areas like active and reinforcement learning, as well as Bayesian optimization and numerical tasks like marginalization in graphical models, the basic challenge is that the algorithm should in some sense ``probe'' an input domain in a maximally informative way, while also ensuring that no area is ignored indefinitely.

As we will review in \textsection\ref{sub:determinantal_point_processes} below, DPPs have a direct connection to the variance function of Gaussian process regression models, which closely ties them to the basic probabilistic algorithms in the aforementioned fields. Of particular interest in the context of Bayesian optimization is the extensive work of \citet{hough2009zeros}, which links the zero-crossings (e.g., the roots of gradients) of a particular class of Gaussian processes to DPPs. In the area of integration---in particular, marginalization in probabilistic models---there is already growing interest in DPPs. The two observations that independent (Monte Carlo) sampling leads to sub-optimal convergence rates of integration estimators \citep{o1987monte}, and that classic quadrature rules can be explicitly interpreted as Gaussian process regression \citep{diaconis88:_bayes,o1991bayes}, has sparked curiosity about structured kernel models for Bayesian quadrature~\citep{osborne2012bayesian,NIPS2012_4657,NIPS2014_5483,NIPS2015_5749}. DPPs have been suggested as exploration strategies for such models, an idea corroborated by a recent theoretical analysis by \citet{2016arXiv160500361B}, who show improved convergence rates for DPP integral estimators compared to Monte Carlo. We discuss this further in \textsection\ref{sub:sampling_from_dpps}, where we also reinterpret an adaptive strategy for rapid Bayesian quadrature by \citet{NIPS2014_5483} in terms of DPPs.

Despite these theoretical arguments in favor of DPPs in continuous spaces, this model class has found only limited practical use as a tool in machine learning and statistics, and almost entirely in discrete domains \citep[e.g.,][]{kulesza2012determinantal,NIPS2013_5008}. One hurdle is that the algorithms typically proposed for sampling DPPs are either computationally taxing or approximate, in particular if the input domain is continuous and/or high-dimensional (more in \textsection\ref{sub:sampling_from_dpps} below). 
The goal of this paper is to point out that this is a historical legacy that should not hold back the use of DPPs as an algorithmic ingredient in machine learning: the historical uses of DPPs, primarily in physics, have model structure that makes exact and efficient sampling difficult. In machine learning, where the designer enjoys algebraic freedom to design the model, this is much less of a problem.
On the other hand, numerical uses like the above often involve a comparably small sample set anyway for outside reasons, which makes inverting Gram matrices an acceptable cost. In \textsection\ref{sec:method}, we present an analytic and exact sampling scheme for DPPs, by way of example on the most popular (Gaussian) kernel in machine learning (generalized in \textsection\ref{sub:other_analytical_kernels}). 

\section{Determinantal point processes}
\label{sub:determinantal_point_processes}

Let $k\colon \mathbb{X}\times\mathbb{X}\to\Re$ be a symmetric positive semi-definite kernel over some bounded space $\mathbb{X}$. We will use a notation familiar to readers of the kernel and Gaussian process literature. Given two sets $A\ce[a_1,\dots,a_I],B\ce[b_1,\dots,b_J]\subseteq \mathbb{X}$, the symbol $K_{AB}\in\Re^{I\times J}$ is a matrix containing the elements $[K_{AB}]_{ij} = k(a_i,b_j)$. For our purposes, a determinantal point process is a stochastic process, such that a finite sample $X\ce[x_1,\dots,x_N] \subset \mathbb{X}$ from the process has joint probability
\begin{equation}
  \label{eq:def_DPP}
  p(x_1,\dots,x_N) = \frac{Z}{N!} \det K_{XX}.
\end{equation}
 Here, $Z$ denotes a normalization constant, the value of which is of minor importance (its existence can be shown using a general argument \citep{hough2006determinantal}). More precise definitions can be found in \citet{soshnikov2000determinantal,hough2009zeros}, and \citet{2016arXiv160500361B}. They require a discussion of base measures and other properties of point processes, which unnecessarily complicate the exposition in our context. \citet{kulesza2012determinantal} also provide a relatively complete introduction to discrete DPPs, where $\mathbb{X}$ is restricted to be a discrete space. We focus on the continuous case.

We make two observations in passing, which may be helpful for intuition. First, Eq.~\eqref{eq:def_DPP} clearly assigns zero measure to samples in which any two distinct samples have the same value, $x_i=x_j,j\neq i$. For such samples, there would be a permutation of $K_{XX}$ that contains at least one $2\times 2$ minor containing exclusively $k(x_i,x_i)$, thus the determinant would vanish. Second, speaking somewhat informally, uniform random point processes can be seen as a special case of DPPs for the Dirac kernel $k(x,x')=\delta(x-x')$. (There is a technical complication since even this kernel does not allow for pairwise identical samples, but such events also have measure zero under random sampling.)

\subsection{Sampling from DPPs --- the connection to active exploration}
\label{sub:sampling_from_dpps}

Equation~\eqref{eq:def_DPP} is an unwieldy object from an algorithmic perspective because it directly addresses the joint distribution of $N$ samples. To iteratively draw samples of increasing size, the conditional distribution $p(x_i\g X_{1:i-1})$ is required, where $X_{1:i-1}\ce [x_1,\dots,x_{i-1}]$. By repeatedly applying the well-known determinant lemma $\det(Z + UWV\Trans) = \det(Z)\det(W)\det(W^{-1}+V\Trans Z^{-1}U)$, this conditional distribution can be found to be~\citep[][Prop.~19]{hough2006determinantal}
\begin{equation}
  \label{eq:iterative_prob}
  p(x_1,\dots,x_N) = Z \prod_{i=1} ^N p(x_i\g X_{1:i-1}) = Z \prod_{i=1} ^N \frac{1}{N-i+1} \V_i(x_i),
\end{equation}
using the function~\cite[see also][]{2016arXiv160500361B}
\begin{equation}
  \label{eq:V}
  \V_i(x) \ce \begin{cases}
  K_{xx} & \text{if } i=1;\\
  K_{xx} - K_{xX_{1:i-1}} K^{-1} _{X_{1:i-1}X_{1:i-1}} K_{X_{1:i-1}x} &\text{otherwise.}
  \end{cases}
\end{equation}
This conditional function will be familiar to readers experienced with Gaussian process models: it is equal to the \emph{posterior predictive variance} of a Gaussian process regression model conditioned on function values at $X_{1:i-1}$. In preparation for the derivations in \textsection\ref{sec:method}, we introduce the shorthand $K_{(i)} \ce K_{1:i-1,1:i-1}$ and re-formulate Eq.~\eqref{eq:V} for $i>1$ more explicitly as
\begin{equation}
  \label{eq:trace}
  \V_i(x) = k(x,x) - \sum_{a,b=1} ^{i-1} k(x,x_a) k(x,x_b) [K_{(i)} ^{-1}]_{ab}.
\end{equation}

Because $\V_i(x)$ can be interpreted as the posterior variance of a Gaussian process model, one can think of a DPP as the point process arising from the following elementary \emph{active learning} strategy: Consider an algorithm aiming to learn the function $f\colon\mathbb{X}\to\Re$ by choosing evaluation points (``designs'') $X$, using a Gaussian process prior $p(f)=\GP(\mu,k)$ with arbitrary mean function $\mu:\mathbb{X}\to\Re$. Aiming to collect informative observations, the algorithm may adopt the policy to evaluate $f$ at a point $x_i$ with probability proportional to $\V_i(x_i)$. If it does so, its designs are samples from the DPP associated with $k$. This strategy can be motivated from within the Bayesian inference framework: evaluating a GP-distributed function at samples from the associated DPP amounts to drawing evaluation points with probability proportional to the expected information gain (evaluating at a mode of the DPP maximizes expected information gain about $f$). 
From outside the Bayesian framework, \citet{2016arXiv160500361B} show that this policy is useful in so far as the resulting empirical estimator $\Exp_N[f]$ for expectations of $f$ (even if $f$ is \emph{not} a true sample from $\GP(\mu,k)$, or even an element of the RKHS associated with $k$) converges at a rate dominating that of the Monte Carlo estimator.

In fact DPPs have already been used, albeit implicitly, in quadrature for machine learning: \citet{NIPS2014_5483} proposed a fast method for Bayesian quadrature to estimate marginal likelihoods. To model the integrand as a strictly positive function, the square root of the likelihood was modeled with a GP. Given data, a second GP was fit to the likelihood, accounting for the nonlinear transformation via linearization or moment matching. This induced GP was used to estimate the desired integral via Bayesian quadrature, \emph{and to choose evaluation points, by (approximately) maximizing the posterior variance of the transformed GP}. The WSABI-L method proposed in the op.cit.~can be seen as choosing MAP samples from a DPP measure. Following \citet{kulesza2012determinantal}, the DPP kernel can be decomposed into a point-wise ``quality'' term and a normalized ``diversity'' kernel.  Setting the former to the posterior mean of the underlying GP, and the latter to the posterior covariance, recovers the model of \citet{NIPS2014_5483}.

From a computational perspective, Eq.~\eqref{eq:iterative_prob} poses two challenges, which seem to have tempered interest in sampling DPPs in continuous spaces, and numerical applications, so far:
\begin{enumerate}
  \item For general kernels $k$, there is usually no analytic cumulative density function for $p(x_i\g X_{1:i-1})$, which is required to draw exact samples from this distribution. Even elaborate studies of sampling methods from DPPs in continuous spaces \citep{scardicchio2009statistical,2016arXiv160500361B} thus rely on rejection sampling---at least for multidimensional domains---sometimes with carefully crafted proposal densities. This problem is pertinent in areas like physics where the kernel $k$ is predetermined by the problem of interest, and tends to have significant structure. This problem is much less severe in machine learning, because our community enjoys freedom in the design of models and can thus choose kernels with convenient analytic properties. Doing so directly yields an exact, efficient algorithm for the generation of samples from DPPs, even in high-dimensional domains $\mathbb{X}$.
  \item Even if the kernel is analytically convenient, Eq.~\eqref{eq:V} involves the matrix inverse of $K_{(i)}$. Given the inverse of $K_{(i-1)}$ from the preceding step in the iterative sampling scheme, this inverse can be computed with complexity $\mathcal{O}\bigl((i-1)^2\bigr)$, using the matrix inversion lemma. Even so, the cost of drawing $N$ samples remains $\mathcal{O}(N^3)$. This issue is directly connected to inference in Gaussian process regression models, and many approximation schemes have been proposed in that area over the past decade. \citet{NIPS2013_4916} proposed leveraging such fast approximation schemes to produce approximate DPP samples in $\mathcal{O}(N)$. We will show in \textsection\ref{sub:degenerate_kernels_and_approximations} that this can introduce significant artifacts. In use cases like Bayesian optimization and quadrature, were the number $N$ of function evaluations is often low, and a Gram matrix is computed/inverted anyway, the cubic cost of exact sampling can be unproblematic, and precision may be more important.
\end{enumerate}

\section{Method}
\label{sec:method}

We point out an algorithm computing exact samples from a DPP if the kernel $k$ is analytically integrable. To ease intuition, the derivations will be by way of example, using the exceedingly popular square-exponential (aka.~Gaussian, RBF) kernel $k_\text{SE}\colon\Re^D\times\Re^D\to\Re$ over the real vector space\footnote{There is a more-general version of this kernel using the Mahalanobis distance induced by a positive definite matrix $\Lambda\in\Re^{D\times D}$. The algorithm as described below can be generalized to that form by rotating the input dimensions to the eigenvectors of $\Lambda$. The kernel $k$ can be scaled by an arbitrary positive scalar $\theta\in\Re_+$, i.e., $k(x,x')\mapsto\theta k(x,x')$ without changing the DPP measure---this simply scales the normalization constant $Z$.}
\begin{equation}
  \label{eq:SE}
  k_\text{SE}(a,b) = \exp\left(- \frac{1}{2} \sum_{d=1} ^D \frac{(a-b)_{d} ^2}{\lambda_d ^2} \right).
\end{equation}
Assume that the sampling domain is the unit cube $x\in[0,1]^D$ (for a more general box constraint $\tilde{x}_d\in[a_d,b_d]$ for each $d=1,\dots,D$, consider the linear transformation $x_d=\nicefrac{\tilde{x_d}-a_d}{b_d-a_d}$). To simplify things even further, we initially consider the univariate problem, $D=1$, then generalize to arbitrary dimensionality. The resulting algorithm draws $N$ samples at cost $\mathcal{O}(DN^2 + N^3)$. A general form of the algorithm is summarized in pseudo-code in Algorithm~\ref{algorithm}.

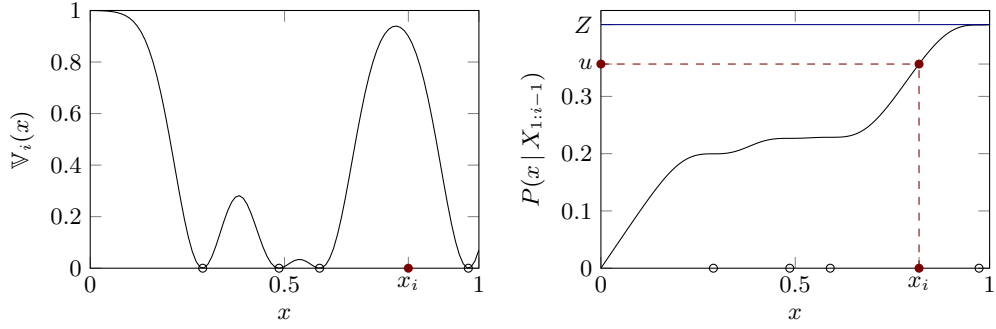
\begin{figure}[t]
  \centering \footnotesize
  \setlength{\figwidth}{0.9\textwidth}
  \setlength{\figheight}{0.15\textheight}
  \input{fig/Drawing_1D}
  \caption{Sketch illustrating analytic sampling from a DPP in one dimension, using the square-exponential kernel~(Eq.~\ref{eq:SE}). Left: conditional probability $\V_i(x)$ for $i=5$. Previous samples $X_{1:4}$ are drawn as empty circles; the new sample is shown in solid red. Right: The sample is drawn by computing the cumulative density $P$ (black line), drawing a scaled uniform random sample $u$ and finding the point $x_i$ such that $P(x_i)=u$, by interval bisection (Alg.~\ref{algorithm}, lines 9--12).}
  \label{fig:sampling_1d}
\end{figure}

\subsection{Sampling in one dimension} 
\label{sub:sampling_in_one_dimension}

\begin{figure}
  \centering \footnotesize
  \setlength{\figwidth}{0.9\textwidth}
  \setlength{\figheight}{0.225\textheight}
  \input{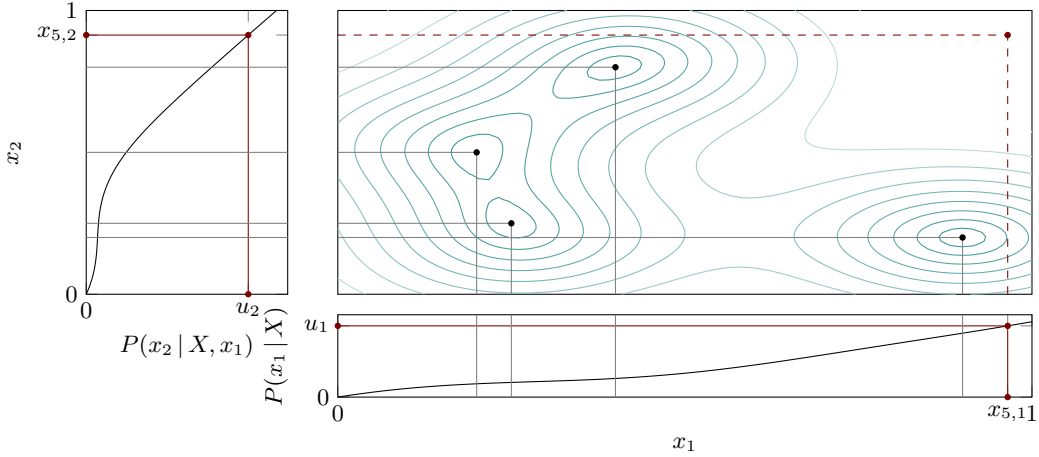}
  \caption{Drawing samples from a DPP in two dimensions (an example sufficiently general to extend to arbitrary dimensionality). Top right: surface plot of the multivariate probability density $\V_5(x)$. The preceding four samples are shown as black points. Bottom: The first coordinate of the fifth sample is drawn first, by computing the marginal density along this dimension. Left: The second element is then drawn by computing a cumulative density \emph{conditioned} on the value of the first element.}
  \label{fig:sampling_2d}
\end{figure}

Exact samples can be drawn using the classic form of computing a non-normalized cumulative density function
\begin{equation}
  \label{eq:CDF_1D}
  P(x\g X_{1:i-1}) = \int_{0} ^x p(x \g X_{1:i-1}) \,\mathrm{d}x,
\end{equation}
and transforming standard uniform random variables $u \sim \mathbb{U}[0,P(1)]$, produced by a pseudo-random number generator, into exact samples from $p$, by setting (cf.~Figure~\ref{fig:sampling_1d})
\begin{equation}
  x = P^{-1}(u \g X_{1:i-1}) = \{x \g P(x) = u \}.
\end{equation}
For the univariate square-exponential kernel~(Eq.~\ref{eq:SE}), Eq.~\eqref{eq:trace} can be re-written, using standard properties of the Gaussian function, as
\begin{equation}
  \label{eq:V_SE_1D}
  \V_i(x) = 1 - \sum_{a,b=1} ^{i-1} \exp\left(- \frac{(x-m_{ab})^2}{\lambda^2} \right) \underbrace{\exp\left(-\frac{(x_a - x_b)^2}{4\lambda^2} \right)}_{\ec M_{(i),ab}} [K_{(i)} ^{-1}]_{ab}.
\end{equation}
where $m_{ab} \ce \nicefrac{1}{2}(x_a + x_b)$, and we have defined a matrix $M_{(i)}\in\Re^{(i-1)\times(i-1)}$. Together, the variables $m,M,K^{-1}$ provide the ``sufficient statistics'' of the sample needed to draw the subsequent point. After $x_i$ has been drawn, these three variables can be updated in $\mathcal{O}(N^2)$---using the matrix inversion lemma to update $K^{-1} _{(i+1)}$; the other two variables can be updated in $\mathcal{O}(N)$. With these objects, the cumulative density is
\begin{equation}
  P(x\g X_{1:i-1}) = x - \frac{\sqrt{\pi}\lambda}{2} \sum_{a,b=1} ^{i-1} \left[\erf\left(\frac{x-m_{ab}}{\lambda} \right) + \erf\biggl(\frac{m_{ab}}{\lambda} \biggr)\right] [M_{(i)} \odot K_{(i)} ^{-1}]_{ab}.
\end{equation}
Here, $\odot$ is the Hadamard (element-wise) product, and we have used $\erf(x)=-\erf(-x)$. Given a uniform random draw $u$, all that is left to do is to find $x$ such that $P(x\g X_{1:i-1})=u$. A straightforward, numerically robust, albeit not particularly ingenious way to do so is by interval bisection. A more elegant search strategy could be constructed using grid refinement methods similar to the popular Ziggurat algorithm of~\citet{marsaglia2000ziggurat}, which we skip here since our goal is merely to highlight generalizable structure, not to find an extremely efficient solution.

\subsection{Multivariate samples}
\label{sub:multivariate_samples}

For square-exponential kernel DPPs in dimension $D>1$, the function $\V_i(x)$ retains much of its structure. Equation~\eqref{eq:V_SE_1D} simply turns into (defining the elements of a new matrix $\mathbf{M}\in\Re^{(i-1)\times(i-1)}$ analogous to $M$ in Eq.~\eqref{eq:V_SE_1D})
\begin{equation}
  \label{eq:V_SE_DD}
    \V_i(x) = 1 - \sum_{a,b=1} ^{i-1} \exp\left(- \sum_{d=1} ^D\frac{(x-m_{ab})_d ^2}{\lambda_d ^2} \right) \underbrace{\exp\left(-\sum_{d=1} ^D \frac{(x_a - x_b)_d ^2}{4\lambda_d ^2} \right)  }_{\ec \mathbf{M}_{(i),ab}} [K_{(i)} ^{-1}]_{ab}.
\end{equation}
The additional challenge in this multivariate case is to construct a parametrization of the cumulative density $P$. This step, too, can be performed in an iterative fashion, drawing one coordinate of the sample point $x_i$ after another (cf.~Figure~\ref{fig:sampling_2d}). Given that the first $d-1$ elements of $x_i$ are given by $x_{i,1:d-1}$, the cumulative density associated with the $d$th dimension is given by the sum rule:
\begin{equation}\label{eq:Multi-D-CDF_integral}
  P(x_{i,d}\g X_{1:i-1},x_{i,1:d-1}) = \int_0 ^{x_{i,d}} \idotsint_0 ^1 p\bigl([x_{i,1:d-1},\tilde{x}_{i,d},\tilde{x}_{i,d+1:}\g X_{1:i-1}\bigr) \,\mathrm{d}\tilde{x}_{i,d} \prod_{\tilde{d}=d+1} ^D \mathrm{d}\tilde{x}_{i,\tilde d}.
\end{equation}
For the square-exponential kernel, this works out to
\begin{equation}
\label{eq:SE_full_P}
\begin{split}
  P([x_i]_d\g X_{1:i-1},x_{i,1:d-1}) &= [x_i]_d - \sum_{a,b=1} ^{i-1} \Bigg\{\exp\left(-\sum_{r=1}^{d-1} \frac{[x_i-m_{ab}]_r ^2}{\lambda_r ^2} \right)
  [\mathbf{M}_{(i)} \odot K_{(i)} ^{-1}]_{ab}\\
  &\q\cdot \left(\erf\left(\frac{[x_i-m_{ab}]_d}{\lambda _d} \right) + \erf\left(\frac{[m_{ab}]_d}{\lambda_d} \right) \right)  \frac{\sqrt{\pi}\lambda_d}{2} \\
  &\q\cdot\left(\prod_{\ell=d+1} ^D \left(\erf\left(\frac{[1-m_{ab}]_\ell}{\lambda _\ell} \right) + \erf\left(\frac{[m_{ab}]_\ell}{\lambda_\ell} \right) \right)  \frac{\sqrt{\pi}\lambda_\ell}{2} \right) \Bigg\}.
\end{split}
\end{equation}
Algorithm~\ref{algorithm} provides a pseudo-code summary.

\begin{algorithm}[t]
\caption{Exact sampling from DPPs with analytic kernels, on $[0,1]^D$.}
\begin{algorithmic}[1]
\Procedure{DrawFromDPP}{$k,D,N$}
\LState $X \gets \varnothing, m\gets\varnothing, M \gets \varnothing, K^{-1} \gets\varnothing$ \Comment{initialize statistics of sample as empty}
\For{$n = 1,\dots,N$} \Comment{draw samples iteratively}
\LState $x_n \gets \varnothing$ \Comment{initialize current sample point}
\For{$d=1,\dots,D$} \Comment{draw dimensions iteratively}
\LState $P\gets$\Call{Pconstruct}{$m,M,K^{-1}$} \Comment{construct function for Eq.~\eqref{eq:SE_full_P}}
\LState $\phantom{P}\mathllap{u}\gets P(1) \cdot$\Call{rand}{$\cdot$} \Comment{draw scaled unit random number}
\LState $\phantom{P}\mathllap{I}\gets[0,1]$ \Comment{initialize search interval}
\While{$|I|>\varepsilon$} \Comment{bisection search}
\LState $\mu \gets \nicefrac{1}{2}(I_0+I_1)$  \Comment{interval midpoint}
\LState $\phantom{\mu}\mathllap{I}\gets$ \textbf{if} $(P(\mu) < u)$ \textbf{then} $[\mu,I_1]$ \textbf{else} {$[I_0,\mu]$} \Comment{bisect}
\EndWhile
\LState $x_n \gets [x_n,\mu]$ \Comment{store sampled element}
\EndFor
\LState $(X,m,M,K^{-1})\gets$\Call{UpdateStats}{$x_n,X,m,M,K^{-1}$} \Comment{update sample statistics (rank-1)}
\EndFor
\EndProcedure
\end{algorithmic}
\label{algorithm}
\end{algorithm}

\subsection{Relation to previous work}
\label{par:relation_to_previous_work}

The main goal of this paper is to point out that applications of DPPs in machine learning and numerics, thanks to their freedom to choose analytically convenient kernels, can utilize exact, computationally efficient sampling schemes in virtually arbitrary dimensionality. The algorithm presented here can be seen as a concrete realization of an abstract recipe introduced by~\citet[][Prop.~19]{hough2006determinantal}.  Other literature on sampling discrete DPPs can be traced back to this work, including \citep{kulesza2012determinantal,NIPS2013_5008,mra_sampling}.  This recipe was investigated numerically by~\citet{scardicchio2009statistical}. However, these authors, restricted by the algebraic structure of their physical application, could only draw analytic samples in 1D and had to resort to rejection sampling in the multivariate case. Although rejection sampling with a decent proposal distribution can scale up to several dimensions, its computational cost rises exponentially with dimensionality. For the high-dimensional domains typical of machine learning problems, only exact sampling is practical.

The work most-related to ours is by \citet{NIPS2013_4916}, who considered \emph{approximate} sampling of DPPs in continuous domains. Here we provide an exact algorithm that should be of interest for Bayesian optimization and integration. We explicitly compare between the quality of exact and approximate samples on a continuous domain \textsection\ref{sub:degenerate_kernels_and_approximations} below.

\subsection{Other analytical kernels}
\label{sub:other_analytical_kernels}

While the Gaussian kernel is the most widely used kernel in machine learning, it has some shortcomings, primarily that it makes very strong smoothness assumptions that can lead to instability in interpolation models. But with some algebraic elbow grease, the scheme of Eq.~\eqref{eq:Multi-D-CDF_integral} can be extended to many other popular kernels, assuming they factorize,
\begin{equation}\label{eq:factorization}
  k(a,b) = \prod_d ^D k(a_d,b_d),
\end{equation}
and the indefinite integrals
\begin{equation}
  \label{eq:necessary_integrals}
  \int  k(a,a)\,\mathrm{d}a \qq\text{and}\qq \int k(a,b)k(a,c) \, \mathrm{d}a
\end{equation}
are analytically solvable. For example, the above results are applicable to the Mat\'{e}rn class of kernels~\citep{stein1999interpolation} (including the exponential kernel, which induces the Ornstein-Uhlenbeck process), noting that, assuming w.l.o.g.~$x_0<a<b<x_1$,
\begin{equation}
  \int_{x_0} ^{x_1} \exp\left(-|x-a|\right)\exp\left(-|x-b|\right) \,\mathrm{d}x = \frac{e^{-a-b}}{2} \left(e^{2a}-e^{2x_0} \right) + (b-a)e^{a-b} + \frac{e^{a+b}}{2}\left(e^{2x_1} - e^{2b} \right),
\end{equation}
and using results such as~\citep[see e.g.,][\textsection 2.322]{gradshteyn2007}
\begin{equation}
  \int x e^{ax}\,\mathrm{d}x = e^{ax}\left(\frac{x}{a}-\frac{1}{a^2}\right),\q \int x^2 e^{ax}\,\mathrm{d}x = e^{ax}\left(\frac{x^2}{a}-\frac{2x}{a^2} + \frac{2}{a^3}\right), \dots
\end{equation}

\section{Comparison to finite-rank approximations}
\label{sub:degenerate_kernels_and_approximations}

\begin{figure}[t]
  \centering\footnotesize
  \input{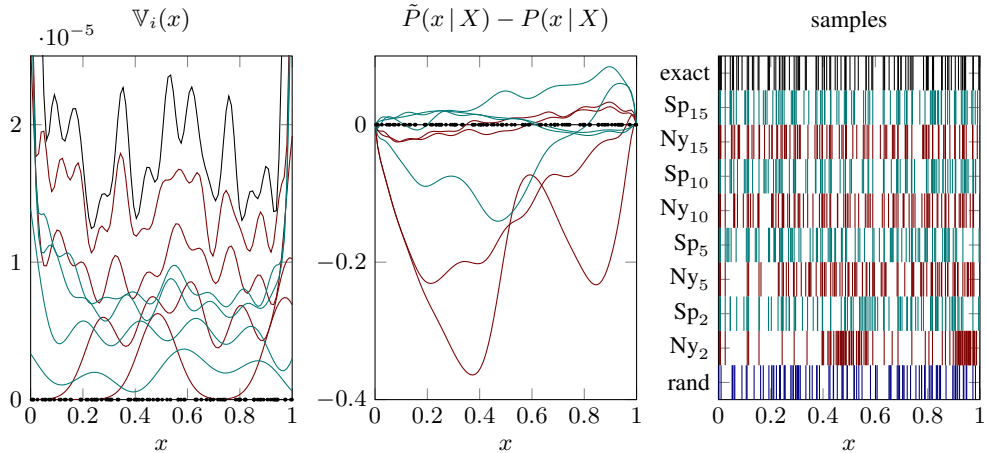}
  \caption{Comparison between exact and approximate sampling schemes. Left: probability density $\V_i(x)$ after $i=100$ exact samples (black dots on abscissa). Exact density in black on top, approximations for Nystr\"om (red) and spectral (green) approximations with $F=5,10,15$ (ordering is obvious, quality increases with $F$). Center: Deviation of the associated approximate cumulative densities $
  \tilde{P}$ from the exact $P$. For this plot, all $\tilde{P}$ and $P$ were normalized (i.e.~$\tilde{P}(1)=1$) for easier comparison, and to clarify that the absolute value of $\V$ is less important than its shape. Right: Sample locations drawn exactly (black, top), and from the approximations. The different samplers used the same random seeds $u$ (shown for comparison in blue at the bottom), so if the approximations were exact, the black and red lines would match. All methods share the first 20 (exact) samples.}
  \label{fig:approximations}
\end{figure}

Due to the matrix inverse $K_{(i)} ^{-1}$ in Eq.~\eqref{eq:trace}, the cost of evaluating $\V_N(x)$ grows cubically, $\mathcal{O}(N^3)$, with the sample size $N$. If large samples are required, an approximate approach of $\mathcal{O}(N)$ may be more appealing. In 2013, \citeauthor{NIPS2013_4916} proposed the use of standard low-rank approximations of the kernel for this purpose. We briefly review this idea here, and compare it empirically to the exact sampler below. If the kernel can be approximated well by a finite-rank expansion
\begin{equation}
  k(a,b) \approx \sigma^2\Id + \sum_{f,g=1} ^F \phi_f (a) \Sigma_{fg} \phi_g(b),
\end{equation}
using a collection of (not necessarily orthogonal) feature functions $\phi\colon\mathbb{X}\to\Re$ and a symmetric positive definite matrix $\Sigma\in\Re^{F\times F}$, then $\V$ can be approximated in $\mathcal{O}(F^3 + NF^2)$ time (i.e., linear in the sample size) as
\begin{equation}
  \V(x) = \sum_{f,g=1} ^F \phi_f(x)\phi_g(x) \left[\Sigma^{-1} + \sigma^{-1} \Phi\Phi\Trans \right]_{fg},
\end{equation}
using the matrix $\Phi\in\Re^{F\times N}$ with elements $\Phi_{fi} = \phi_f(x_i)$. This includes the case of degenerate kernels, i.e., where $k$ is \emph{exactly} captured by such a finite-rank expansion, such as in simple linear and polynomial regression. Approximate cases include the Nystr\"om approximation \citep{NIPS2000_1866}, and spectral expansions~\citep{NIPS2007_3182}. The latter approach works for any translation invariant kernel over the unit hypercube (including the square-exponential), since the necessary eigenfunctions are the trigonometric functions. The necessary integrals in Eq.~\eqref{eq:necessary_integrals} can be solved using identities like~\citep[cf.][\textsection 2.532, for a more complete list]{gradshteyn2007}
\begin{alignat}{2}
    \int \cos(ax)\cos(bx) \,\mathrm{d}x &\operatorname*{=}^{a\neq b} +\frac{\sin((a-b)x)}{2(a-b)} + \frac{\sin((a+b)x)}{2(a+b)} &\q\text{or} &\operatorname*{=}^{a=b} \frac{x}{2} + \frac{\sin(2ax)}{4a}.
\end{alignat}
Figure \ref{fig:approximations} empirically compares samples drawn from a univariate DPP using these two approximations with exact samples drawn using the algorithm outlined above. It shows both the approximations to the density $\V$, the differences of the induced normalized cumulative density functions to the exact one, and approximate DPP samples, for both the Nystr\"om and spectral approximations of varying fidelity $F$ (further details in caption). Rough approximations (i.e. with small $F$) can perform badly---in fact, the samples drawn this way may be \emph{more} clumped than uniform random samples. This experiment was deliberately performed in 1D to take an optimistic stance on the potential quality of the approximations: For DPP sampling, the number of features necessary to cover higher-dimensional spaces increases exponentially with dimension. (Empirical evaluations of low-rank approximations for kernel regression show that they can work well even in multivariate settings~\citep[e.g.,][]{NIPS2007_3182}, but empirical datasets tend to lie on low-dimensional manifolds, while DPP samples, by construction, cover the entire domain, thus require the approximation to actually cover the space.)

There is thus a nontrivial trade-off between cost and precision when sampling from a DPP. If a moderate number of samples (say, $N\lesssim10^3$) are to be drawn in a space of high dimension $D$, it may be a better idea to draw exact samples at cost $\O(N^3 + N^2D)$, rather than approximate samples to rank $F$ at cost $\O(NF^2 + F^3 + F^2 D)$, if $F$ scales exponentially with $D$. This is particularly true if, as in Bayesian optimization and quadrature, the reason to draw the samples is to condition a GP regression model that already requires the very same cubic cost matrix inversion anyway.

\section{Conclusion}
\label{sec:conclusion}

We have pointed out that, for popular kernel classes of machine learning and statistics, it is often possible to analytically draw exact samples from determinantal point processes on continuous spaces at comparably low computational cost. More precisely, this is true if the kernel defining the DPP has certain analytic properties defined in Eqs.~\eqref{eq:factorization} \&~\eqref{eq:necessary_integrals}. If the resulting sample locations are used to condition a kernel regressor, the sampling algorithm's cost is dominated by a matrix inversion that is required by the regressor anyway, with a very small remaining cost overhead. This immediate association between a kernel regressors and a DPP, coupled with the interesting theoretic properties of DPPs, make these point processes an interesting framework for areas like Bayesian optimization and quadrature. The algorithm outlined above provides the necessary tool to utilize this opportunity.

\subsection*{Acknowledgements}

The authors are grateful to Lucy Kuncheva and Joseph Courtney for (separately) pointing out a nontrivial typo in Eq.~\eqref{eq:SE_full_P} in an earlier version of this manuscript.

\small
\bibliographystyle{abbrvnat}
\bibliography{bibfile}

\end{document}

%% file: preamble.tex
\usepackage{microtype,marvosym} 

\usepackage{tikz,pgfplots}
\pgfplotsset{compat=newest}
\pgfplotsset{plot coordinates/math parser=false}       

\definecolor{lred}{RGB}{200,0,0}
\definecolor{dred}{RGB}{130,0,0} \definecolor{dblu}{RGB}{0,0,130}
\definecolor{dgre}{RGB}{0,130,0} \definecolor{dgra}{RGB}{50,50,50}
\definecolor{mgra}{RGB}{100,100,100}
\definecolor{lgra}{RGB}{220,220,220}
\definecolor{MPG}{RGB}{000,125,122}
\definecolor{ora}{HTML}{FF9933}

\definecolor{AMPurple}{HTML}{663366}
\definecolor{Burgundy}{HTML}{993333}
\definecolor{Coffee}{HTML}{7B6049}
\definecolor{ForestGreen}{HTML}{005826}
\definecolor{Lavender}{HTML}{6E6AB1}
\definecolor{PSLightBlue}{HTML}{7DA7D9}

\newcommand{\g}{\,|\,}

\newcommand{\Exp}{\mathbb{E}}

\newcommand{\erf}{\operatorname{erf}}

\renewcommand{\Re}{\mathbb{R}}

\newcommand{\Trans}{^{\intercal}}

\renewcommand{\det}{\operatorname{det}}

\newcommand{\q}{\quad}
\newcommand{\qq}{\qquad}

\renewcommand{\vec}{\boldsymbol}

\renewcommand{\O}{\mathcal{O}} 
 
\newcommand{\GP}{\mathcal{GP}}
\newcommand{\Id}{\vec{I}}

\newcommand{\V}{\mathbb{V}}

\usepackage{colonequals}
\newcommand{\ce}{\colonequals}
\newcommand{\ec}{\equalscolon}

\usetikzlibrary{arrows,shapes,plotmarks}

\tikzset{>=stealth'} 
\tikzstyle{graphnode} = 
   [circle,draw=black,minimum size=22pt,text centered,text
     width=22pt,inner sep=0pt] 
\tikzstyle{var}   =[graphnode,fill=white]
\tikzstyle{obs}   =[graphnode,fill=black,text=white]
\tikzstyle{fac}   =[rectangle,draw=black,fill=black!25,minimum size=5pt]
\tikzstyle{facprior} =[rectangle,draw=black,fill=black,text=white,minimum size=5pt]
\tikzstyle{edge}  =[draw=white,double=black,thick,-]
\tikzstyle{prior} =[rectangle, draw=black, fill=black, minimum size=
5pt, inner sep=0pt]
\tikzstyle{dirprior} = [circle, draw=black, fill=black, minimum
size=5pt, inner sep=0pt]

\DeclareSymbolFont{stmry}{U}{stmry}{m}{n}
\DeclareMathSymbol\leftarrowtriangle\mathrel{stmry}{"5E}
\DeclareMathSymbol\rightarrowtriangle\mathrel{stmry}{"5F}
\DeclareMathSymbol\leftrightarrowtriangle\mathrel{stmry}{"5D}
\DeclareMathSymbol\obar\mathrel{stmry}{"3A}
\DeclareMathSymbol\otimes\mathrel{stmry}{"0F}
\DeclareMathSymbol\ominus\mathrel{stmry}{"17}
\DeclareMathSymbol\sslash\mathrel{stmry}{"0C}
\renewcommand{\gets}{\operatorname*{\leftarrowtriangle}}
\renewcommand{\to}{\operatorname*{\rightarrowtriangle}}

\usepackage{mathtools}
\usepackage{algorithm}
\usepackage{algpseudocode}
\algrenewcommand{\algorithmiccomment}[1]{\hfill {\footnotesize $\sslash$ #1}}
\algrenewcommand{\alglinenumber}[1]{\sf\scriptsize #1}

\makeatletter
\@addtoreset{algorithm}{chapter} 
\makeatother

\makeatletter
\def\therule{\makebox[\algorithmicindent][l]{\hspace*{.5em}\vrule height .75\baselineskip depth .25\baselineskip}}%

\newtoks\therules
\therules={}
\def\appendto#1#2{\expandafter#1\expandafter{\the#1#2}}
\def\gobblefirst#1{
  #1\expandafter\expandafter\expandafter{\expandafter\@gobble\the#1}}%
\def\LState{\State\unskip\the\therules}
\def\pushindent{\appendto\therules\therule}%
\def\popindent{\gobblefirst\therules}%
\def\printindent{\unskip\the\therules}%
\def\printandpush{\printindent\pushindent}%
\def\popandprint{\popindent\printindent}%

\algdef{SE}[WHILE]{While}{EndWhile}[1]
  {\printandpush\algorithmicwhile\ #1\ \algorithmicdo}
  {\popandprint\algorithmicend\ \algorithmicwhile}%
\algdef{SE}[FOR]{For}{EndFor}[1]
  {\printandpush\algorithmicfor\ #1\ \algorithmicdo}
  {\popandprint\algorithmicend\ \algorithmicfor}%
\algdef{S}[FOR]{ForAll}[1]
  {\printindent\algorithmicforall\ #1\ \algorithmicdo}%
\algdef{SE}[LOOP]{Loop}{EndLoop}
  {\printandpush\algorithmicloop}
  {\popandprint\algorithmicend\ \algorithmicloop}%
\algdef{SE}[REPEAT]{Repeat}{Until}
  {\printandpush\algorithmicrepeat}[1]
  {\popandprint\algorithmicuntil\ #1}%
\algdef{SE}[IF]{If}{EndIf}[1]
  {\printandpush\algorithmicif\ #1\ \algorithmicthen}
  {\popandprint\algorithmicend\ \algorithmicif}%
\algdef{C}[IF]{IF}{ElsIf}[1]
  {\popandprint\pushindent\algorithmicelse\ \algorithmicif\ #1\ \algorithmicthen}%
\algdef{Ce}[ELSE]{IF}{Else}{EndIf}
  {\popandprint\pushindent\algorithmicelse}%
\algdef{SE}[PROCEDURE]{Procedure}{EndProcedure}[2]
   {\printandpush\algorithmicprocedure\ \textproc{#1}\ifthenelse{\equal{#2}{}}{}{(#2)}}%
   {\popandprint\algorithmicend\ \algorithmicprocedure}%
\algdef{SE}[FUNCTION]{Function}{EndFunction}[2]
   {\printandpush\algorithmicfunction\ \textproc{#1}\ifthenelse{\equal{#2}{}}{}{(#2)}}%
   {\popandprint\algorithmicend\ \algorithmicfunction}%
\makeatother

%% file: fig/Samples.tex
%
%
\begin{tikzpicture}

\begin{axis}[%
width=0.275\figwidth,
height=0.275\figwidth,
at={(0\figwidth,0\figwidth)},
scale only axis,
xmin=0,
xmax=1,
xtick={0, 1},
xlabel={$x_1$},
ymin=0,
ymax=1,
ytick={0, 1},
ylabel={$x_2$},
axis background/.style={fill=white},
ylabel near ticks,
xlabel near ticks
]
\addplot [color=black,mark size=1.0pt,only marks,mark=*,mark options={solid,fill=black},forget plot]
  table[row sep=crcr]{%
0.944484785122104	0.116644169672164\\
0.851683634304443	0.166714469869819\\
0.695714931502714	0.460396475065048\\
0.718793709053039	0.791177473115077\\
0.244537721280044	0.614089458310341\\
0.45993592399973	0.705708882320399\\
0.340970996546437	0.794444518125157\\
0.502988907808819	0.024210455832259\\
0.473826854364154	0.78697394781355\\
0.558034099630536	0.626582293937794\\
0.450934905289743	0.252868171557484\\
0.448954868819347	0.284418051185094\\
0.98773761854596	0.0735105308502372\\
0.658092077834797	0.824636078759346\\
0.653746435969567	0.457789305486818\\
0.0375517794175529	0.682254396185121\\
0.941557652823287	0.810711020333221\\
0.571130434289421	0.958228262623494\\
0.100633162017525	0.723750401220885\\
0.0189570612943981	0.932027245355523\\
0.504312013157189	0.712847800179654\\
0.630692560101382	0.900003244461972\\
0.964281162649267	0.101196087236093\\
0.832320343521725	0.86176802087041\\
0.538441354630034	0.489213369641999\\
0.463434088413499	0.358017327580542\\
0.661063367564783	0.142176132901745\\
0.818272994616129	0.961557916899145\\
0.262267699701707	0.770939848034392\\
0.512572617612012	0.405115055055923\\
0.0756929327189357	0.43492224693906\\
0.722662313761174	0.64880286887919\\
0.606358926403091	0.389882256426137\\
0.97041003753868	0.224149388494633\\
0.549649879913011	0.299818971232975\\
0.197738487906687	0.304158743707506\\
0.109421093627614	0.452013184347154\\
0.0607555251132625	0.982540481771167\\
0.471401955171776	0.896727927903791\\
0.389406400097177	0.439319482315454\\
0.610221137518591	0.694108318771682\\
0.074445276966626	0.145968838957519\\
0.272086984294932	0.231849344752366\\
0.177727893738966	0.167925903552344\\
0.370167889369021	0.0274016872236911\\
0.305023107878509	0.328705000189688\\
0.933010656689956	0.755569203137927\\
0.375615101790811	0.138347843455466\\
0.414460029261267	0.342304569740014\\
0.319814619488908	0.391055139088597\\
0.903939133675155	0.759009641576747\\
0.0355035362338955	0.839462199221502\\
0.70348076672853	0.985972476668269\\
0.190410385028478	0.479811370360489\\
0.355227532903309	0.957456993715325\\
0.378229011389484	0.562899782100245\\
0.0322419279844685	0.862459443699292\\
0.87532396406443	0.89923951138733\\
0.10905262552208	0.789644012757796\\
0.0294693024163448	0.87097052895573\\
0.383552330825032	0.315454596845608\\
0.198795611176378	0.0467088175915427\\
0.43607070976541	0.875731862668242\\
0.898386670003947	0.105577683412437\\
0.51176292269245	0.373133710294757\\
0.866506853534358	0.939914946506772\\
0.517224178017347	0.811978886825857\\
0.644478279885376	0.453783530880991\\
0.156064342499071	0.910244208813189\\
0.964502036266762	0.711439529308697\\
0.817937982590658	0.97933843248634\\
0.402857815031654	0.374801945429665\\
0.823531737475173	0.213491237037688\\
0.692464977559561	0.922853757838673\\
0.727310714865602	0.296104200017496\\
0.904419465342146	0.315072020890235\\
0.520570946590059	0.41849192665035\\
0.14813460396623	0.375693227464137\\
0.823802861816351	0.81652212677923\\
0.677603738794111	0.606809361820259\\
0.157295006472349	0.394515958067669\\
0.0499102813453245	0.0452691312914888\\
0.378422033605598	0.304025297522915\\
0.178875282873983	0.407016762901019\\
0.25037419604213	0.363256976143107\\
0.865689053205219	0.678033483109644\\
0.692192958688194	0.223141185252334\\
0.843035712536411	0.633947086271661\\
0.270943882929472	0.0769321494882687\\
0.379813633297878	0.0453968851545402\\
0.0258451613565874	0.686069531054907\\
0.239832506038597	0.14602206279002\\
0.936023554751089	0.386165059794558\\
0.845867278243271	0.799507618929964\\
0.846394730694219	0.775696924827323\\
0.310797781443229	0.712612464771046\\
0.474243999914829	0.733539818817262\\
0.850317356026865	0.686782578102685\\
0.499300922694249	0.878999340076414\\
0.203274659585837	0.283417140661536\\
};
\end{axis}

\begin{axis}[%
width=0.275\figwidth,
height=0.275\figwidth,
at={(0.362\figwidth,0\figwidth)},
scale only axis,
xmin=0,
xmax=1,
xtick={0, 1},
xlabel={$x_1$},
ymin=0,
ymax=1,
ytick={0, 1},
axis background/.style={fill=white},
ylabel near ticks,
xlabel near ticks
]
\addplot [color=black,mark size=1.0pt,only marks,mark=*,mark options={solid,fill=black},forget plot]
  table[row sep=crcr]{%
0.990281312726438	0.240078260190785\\
0.893880222924054	0.423128145746887\\
0.554642184637487	0.484422703273594\\
0.560803615488112	0.914861864410341\\
0.75738547835499	0.0408649398013949\\
0.375619740225375	0.295982199721038\\
0.908153061755002	0.984189639799297\\
0.124463767744601	0.615562188439071\\
0.749933854676783	0.165882107801735\\
0.341342237778008	0.917280445806682\\
0.294428600929677	0.100792641751468\\
0.316344038583338	0.817969982512295\\
0.855268496088684	0.0870627919211984\\
0.984816580079496	0.465447877533734\\
0.739447760395706	0.994240649975836\\
0.128110160119832	0.193808785639703\\
0.956705440767109	0.526703597046435\\
0.524088363163173	0.232510478235781\\
0.877466008998454	0.791462280787528\\
0.296117392368615	0.555140775628388\\
0.51632131729275	0.534447920508683\\
0.397248051129282	0.435916162095964\\
0.247278195805848	0.49586873780936\\
0.759163725189865	0.855049493722618\\
0.167353601194918	0.727601523511112\\
0.158318954519928	0.372289317660034\\
0.641400367952883	0.718971024267375\\
0.0814148122444749	0.963310887105763\\
0.992184591479599	0.701372832991183\\
0.855656213127077	0.701627311296761\\
0.233682612888515	0.603527289815247\\
0.343923230655491	0.735011317767203\\
0.335114509798586	0.0128689156845212\\
0.70404979493469	0.411082656122744\\
0.413303381763399	0.674754000268877\\
0.381681284867227	0.111771472729743\\
0.227602214552462	0.306361990980804\\
0.792827499099076	0.411874328739941\\
0.399176876060665	0.631934545002878\\
0.0360098788514733	0.86244483012706\\
0.065917412750423	0.575190407224\\
0.434944898821414	0.33350249286741\\
0.887848087586462	0.527019538916647\\
0.914722268469632	0.0392970079556108\\
0.56235630903393	0.807730228640139\\
0.173853267915547	0.803706555627286\\
0.474300409667194	0.140416690148413\\
0.817639858461916	0.225590162910521\\
0.136991848237813	0.0314607052132487\\
0.736810239963233	0.667410609312356\\
0.556203532032669	0.555731925182045\\
0.00729493331164122	0.643298828043044\\
0.733981898985803	0.341877688653767\\
0.377005499787629	0.903022511862218\\
0.994319002144039	0.127879672683775\\
0.196403955109417	0.917382528074086\\
0.0464271539822221	0.470607570372522\\
0.0381383011117578	0.094407475553453\\
0.948034971021116	0.930901329033077\\
0.0771538550034165	0.321310735307634\\
0.0584068289026618	0.0449554780498147\\
0.906058323569596	0.231710379011929\\
0.223203488625586	0.159794588573277\\
0.57306304294616	0.0629114983603358\\
0.92064571660012	0.329676282592118\\
0.372091650031507	0.503003706224263\\
0.311790429987013	0.986707362346351\\
0.556947411037982	0.217572097666562\\
0.658181712962687	0.92403316590935\\
0.761264673434198	0.475961457006633\\
0.286654784344137	0.37140781339258\\
0.091193332336843	0.402927716262639\\
0.498914266936481	0.754268052987754\\
0.598781402222812	0.329642101190984\\
0.452984659932554	0.821598005481064\\
0.2152388850227	0.949069905094802\\
0.168879103846848	0.633092460222542\\
0.358668700791895	0.0328818829730153\\
0.303206485696137	0.271753010340035\\
0.00111301895231009	0.432909994386137\\
0.859772373922169	0.9355627624318\\
0.862713470123708	0.613336740992963\\
0.600439195521176	0.972263277508318\\
0.191983220167458	0.470251043327153\\
0.684496072120965	0.0702083902433515\\
0.958178407512605	0.798220657743514\\
0.00802145805209875	0.331691476516426\\
0.646454119123518	0.601525674574077\\
0.0435206731781363	0.909298018552363\\
0.676504063419998	0.481428301893175\\
0.431804870255291	0.00719096418470144\\
0.680215687491	0.229587552137673\\
0.982546179555357	0.865339390002191\\
0.994246837683022	0.947034173645079\\
0.403965440578759	0.172420264221728\\
0.736889009363949	0.596228660084307\\
0.922421968542039	0.645696387626231\\
0.46574782859534	0.677799663506448\\
0.615847361274064	0.237973396666348\\
0.113852511160076	0.813534625805914\\
};
\addplot [color=black,solid,line width=1.0pt,forget plot]
  table[row sep=crcr]{%
0.1	0.1\\
0.2	0.1\\
};
\addplot [color=black,solid,line width=1.0pt,forget plot]
  table[row sep=crcr]{%
0.1	0.1\\
0.1	0.2\\
};
\node[right, align=left, text=black]
at (axis cs:0.2,0.1) {$\lambda_1$};
\node[right, align=left, text=black]
at (axis cs:0.1,0.2) {$\lambda_2$};
\end{axis}

\begin{axis}[%
width=0.275\figwidth,
height=0.275\figwidth,
at={(0.725\figwidth,0\figwidth)},
scale only axis,
xmin=0,
xmax=1,
xtick={0, 1},
xlabel={$x_1$},
ymin=0,
ymax=1,
ytick={0, 1},
axis background/.style={fill=white},
ylabel near ticks,
xlabel near ticks
]
\addplot [color=black,mark size=1.0pt,only marks,mark=*,mark options={solid,fill=black},forget plot]
  table[row sep=crcr]{%
0.756409539841115	0.233611074276268\\
0.618400567211211	0.80183604452759\\
0.768451779149473	0.838745485059917\\
0.0874828426167369	0.219190924428403\\
0.524771277792752	0.32026553992182\\
0.356073199771345	0.735237547196448\\
0.0221077231690288	0.348466130904853\\
0.260229439474642	0.00689924973994493\\
0.965282688848674	0.748508782126009\\
0.131907573901117	0.882087153382599\\
0.0718496283516288	0.874518281780183\\
0.618719414807856	0.551702049560845\\
0.888706949539483	0.979976099915802\\
0.0198566047474742	0.671085099689662\\
0.243035362102091	0.979591331444681\\
0.488676988519728	0.565351969562471\\
0.802464549429715	0.038591924123466\\
0.362776325084269	0.978255099616945\\
0.213132788427174	0.091420111246407\\
0.456967779435217	0.162321605719626\\
0.927769492380321	0.392124871723354\\
0.119267356581986	0.449035421945155\\
0.980033026076853	0.236715228296816\\
0.623099361546338	0.994814797304571\\
0.416814084164798	0.414775152690709\\
0.752950930036604	0.639164085499942\\
0.528822806663811	0.967155997641385\\
0.320882036350667	0.476114022545516\\
0.656313522718847	0.169755249284208\\
0.955735142342746	0.17905078176409\\
0.52460344042629	0.631403501145542\\
0.237486871890724	0.840081811882555\\
0.744634778238833	0.0421410007402301\\
0.839264522306621	0.711880403570831\\
0.423354328610003	0.0852011060342193\\
0.471440001390874	0.998848480172455\\
0.889021364040673	0.58484270516783\\
0.0678383810445666	0.670805674977601\\
0.49039760697633	0.0905722687020898\\
0.239300579763949	0.523759483359754\\
0.850807744078338	0.486194443888962\\
0.116648784838617	0.00885068904608488\\
0.950676272623241	0.499728155322373\\
0.452896556816995	0.625161967240274\\
0.864661377854645	0.0388216180726886\\
0.973713190294802	0.047132913954556\\
0.00596984941512346	0.0180638385936618\\
0.354821634478867	0.260357900522649\\
0.755512376315892	0.995393297635019\\
0.684107004664838	0.719462369568646\\
0.000699537806212902	0.694001647643745\\
0.624120286665857	0.0181893957778811\\
0.00131915789097548	0.976112839765847\\
0.168424132280052	0.830483230762184\\
0.323662300594151	0.97209411393851\\
0.567399461753666	0.0744527159258723\\
0.985413580201566	0.742417420260608\\
0.518126499839127	0.835175291635096\\
0.954430685378611	0.95929950941354\\
0.113149776123464	0.0902758920565248\\
0.795883974991739	0.976083165965974\\
0.0448656594380736	0.00134002510458231\\
0.947861247695982	0.114411839284003\\
0.23886742349714	0.290424239821732\\
0.155243362300098	0.980718963779509\\
0.301321598701179	0.0459086922928691\\
0.99898964073509	0.983165257610381\\
0.90824010130018	0.834106800146401\\
0.698090135119855	0.502432306297123\\
0.158377832733095	0.154769937507808\\
0.470768316648901	0.85331507679075\\
0.877855709753931	0.328971520997584\\
0.751865944825113	0.768747807480395\\
0.651633129455149	0.90354794356972\\
0.179245908744633	0.556627386249602\\
0.0665412852540612	0.530938983894885\\
0.455233092419803	0.448604724369943\\
0.50484383944422	0.0173681089654565\\
0.839217365719378	0.23325986135751\\
0.723083705641329	0.560948804952204\\
0.357652918435633	0.000635263510048389\\
0.307426952756941	0.717333539389074\\
0.0845713438466191	0.0592461070045829\\
0.129028090275824	0.758460043929517\\
0.545184372924268	0.155850718729198\\
0.216997249983251	0.524948150850832\\
0.290652868337929	0.279291571117938\\
0.00512078311294317	0.471110939048231\\
0.42353202495724	0.945897801779211\\
0.680986405350268	0.887589042074978\\
0.639942883513868	0.541441894136369\\
0.703123736195266	0.91880991961807\\
9.31322574615479e-10	0.190896849147975\\
9.31322574615479e-10	0.829153061844409\\
9.31322574615479e-10	9.31322574615479e-10\\
0.758662377484143	0.948532135225832\\
0.695233295671642	0.595740835182369\\
9.31322574615479e-10	0.214297248981893\\
9.31322574615479e-10	0.432011931203306\\
9.31322574615479e-10	9.31322574615479e-10\\
};
\addplot [color=black,solid,line width=1.0pt,forget plot]
  table[row sep=crcr]{%
0.1	0.1\\
0.2	0.1\\
};
\addplot [color=black,solid,line width=1.0pt,forget plot]
  table[row sep=crcr]{%
0.1	0.1\\
0.1	0.6\\
};
\node[right, align=left, text=black]
at (axis cs:0.2,0.1) {$\lambda_1$};
\node[right, align=left, text=black]
at (axis cs:0.1,0.6) {$\lambda_2$};
\end{axis}
\end{tikzpicture}%

%% file: fig/Drawing_1D.tex
%
%
\definecolor{mycolor1}{rgb}{0.00000,0.44700,0.74100}%
\definecolor{mycolor2}{rgb}{0.49060,0.00000,0.00000}%
\definecolor{mycolor3}{rgb}{0.00000,0.00000,0.50900}%
\begin{tikzpicture}

\begin{axis}[%
width=0.411\figwidth,
height=\figheight,
at={(0\figwidth,0\figheight)},
scale only axis,
xmin=0,
xmax=1,
xtick={0,0.5,0.818456492386758,1},
xticklabels={{$0$},{$0.5$},{$x_i$},{$1$}},
xlabel={$x$},
ymin=0,
ymax=1,
ylabel={$\mathbb{V}_i(x)$},
axis background/.style={fill=white},
ylabel near ticks,
xlabel near ticks
]
\addplot [color=black,solid,forget plot]
  table[row sep=crcr]{%
0	0.999762585785801\\
0.0101010101010101	0.999578338260902\\
0.0202020202020202	0.999266238111993\\
0.0303030303030303	0.998748938510212\\
0.0404040404040404	0.997910061106426\\
0.0505050505050505	0.996579271657004\\
0.0606060606060606	0.994514300661851\\
0.0707070707070707	0.991380703104155\\
0.0808080808080808	0.986731076183015\\
0.0909090909090909	0.979986608643986\\
0.101010101010101	0.970425096947271\\
0.111111111111111	0.957180708938561\\
0.121212121212121	0.939261484607039\\
0.131313131313131	0.915590451628236\\
0.141414141414141	0.885074921599019\\
0.151515151515152	0.846705749364461\\
0.161616161616162	0.799684037406752\\
0.171717171717172	0.743567239435447\\
0.181818181818182	0.678420546185001\\
0.191919191919192	0.604953875799531\\
0.202020202020202	0.524621031787063\\
0.212121212121212	0.439656922781846\\
0.222222222222222	0.353032140610301\\
0.232323232323232	0.268312027562633\\
0.242424242424242	0.189419145053906\\
0.252525252525253	0.120312388050439\\
0.262626262626263	0.0646107129760606\\
0.272727272727273	0.0252019908904888\\
0.282828282828283	0.00388537599252736\\
0.292929292929293	0.00109691250556176\\
0.303030303030303	0.0157620403405202\\
0.313131313131313	0.0453055921650262\\
0.323232323232323	0.0858313828624134\\
0.333333333333333	0.13246205745618\\
0.343434343434343	0.179808420265461\\
0.353535353535354	0.222518928031904\\
0.363636363636364	0.25584688432676\\
0.373737373737374	0.276166940004878\\
0.383838383838384	0.281374819906195\\
0.393939393939394	0.271114985774031\\
0.404040404040404	0.246799609439078\\
0.414141414141414	0.211407283939694\\
0.424242424242424	0.16907889116838\\
0.434343434343434	0.12455753429077\\
0.444444444444444	0.0825451810648402\\
0.454545454545455	0.0470660647209482\\
0.464646464646465	0.0209318316839111\\
0.474747474747475	0.00539326339820123\\
0.484848484848485	3.79295634602705e-05\\
0.494949494949495	0.00295520040340747\\
0.505050505050505	0.0111455100124769\\
0.515151515151515	0.0211076344079394\\
0.525252525252525	0.0295046349588379\\
0.535353535353535	0.0337932691909894\\
0.545454545454545	0.0327071301641736\\
0.555555555555556	0.0265101755135394\\
0.565656565656566	0.0169796993504213\\
0.575757575757576	0.007127639365893\\
0.585858585858586	0.000716184894902883\\
0.595959595959596	0.00165845949419441\\
0.606060606060606	0.0134110059215732\\
0.616161616161616	0.0384596613528098\\
0.626262626262626	0.0779765939477972\\
0.636363636363636	0.131690009563402\\
0.646464646464647	0.197967659553301\\
0.656565656565657	0.274079139870174\\
0.666666666666667	0.356576569188744\\
0.676767676767677	0.441722001288246\\
0.686868686868687	0.525892892912141\\
0.696969696969697	0.605911250837479\\
0.707070707070707	0.67926307390723\\
0.717171717171717	0.744197282923853\\
0.727272727272727	0.799713117478489\\
0.737373737373737	0.845459108460988\\
0.747474747474748	0.88157412470035\\
0.757575757575758	0.908502145016164\\
0.767676767676768	0.926808900626404\\
0.777777777777778	0.937022385537308\\
0.787878787878788	0.939512324774975\\
0.797979797979798	0.934417365189278\\
0.808080808080808	0.921623646904334\\
0.818181818181818	0.900794502207152\\
0.828282828282828	0.871447841126207\\
0.838383838383838	0.83307469382193\\
0.848484848484849	0.785288908888234\\
0.858585858585859	0.727994050544314\\
0.868686868686869	0.661549483286648\\
0.878787878787879	0.586914321863642\\
0.888888888888889	0.505746472472115\\
0.898989898989899	0.420435489433405\\
0.909090909090909	0.334053141883356\\
0.919191919191919	0.25021446934209\\
0.929292929292929	0.172853885455522\\
0.939393939393939	0.105933907883323\\
0.94949494949495	0.0531161000709873\\
0.95959595959596	0.017432410466044\\
0.96969696969697	0.000998286066391629\\
0.97979797979798	0.00480562580735644\\
0.98989898989899	0.0286239722008022\\
1	0.0710237609479378\\
};
\addplot [color=black,mark size=1.5pt,only marks,mark=o,mark options={solid},forget plot]
  table[row sep=crcr]{%
0.972857342101634	0\\
0.485820201225579	0\\
0.590023984201252	0\\
0.289399405010045	0\\
};
\addplot [color=mycolor1,mark size=1.5pt,only marks,mark=*,mark options={solid,fill=mycolor2,draw=mycolor2},forget plot]
  table[row sep=crcr]{%
0.818456492386758	0\\
};
\end{axis}

\begin{axis}[%
width=0.411\figwidth,
height=\figheight,
at={(0.54\figwidth,0\figheight)},
scale only axis,
xmin=0,
xmax=1,
xtick={0,0.5,0.818456492386758,1},
xticklabels={{$0$},{$0.5$},{$x_i$},{$1$}},
xlabel={$x$},
ymin=0,
ymax=0.45,
ytick={0,0.1,0.2,0.3,0.356490294977174,0.425334483932571},
yticklabels={{$0$},{$0.1$},{$0.2$},{$0.3$},{$u$},{$Z$}},
ylabel={$P(x\g X_{1:i-1})$},
axis background/.style={fill=white},
ylabel near ticks,
xlabel near ticks
]
\addplot [color=black,mark size=1.5pt,only marks,mark=o,mark options={solid},forget plot]
  table[row sep=crcr]{%
0.972857342101634	0\\
0.485820201225579	0\\
0.590023984201252	0\\
0.289399405010045	0\\
};
\addplot [color=black,solid,forget plot]
  table[row sep=crcr]{%
0	0\\
0.0101010101010101	0.0100977645443917\\
0.0202020202020202	0.0201930743273922\\
0.0303030303030303	0.0302842747038388\\
0.0404040404040404	0.0403687438198834\\
0.0505050505050505	0.0504424256900228\\
0.0606060606060606	0.0604991967426022\\
0.0707070707070707	0.0705300383200874\\
0.0808080808080808	0.0805220000909995\\
0.0909090909090909	0.0904569623684608\\
0.101010101010101	0.10031024071638\\
0.111111111111111	0.110049123998461\\
0.121212121212121	0.119631494469131\\
0.131313131313131	0.129004739299509\\
0.141414141414141	0.138105216827007\\
0.151515151515152	0.146858574181404\\
0.161616161616162	0.15518121043837\\
0.171717171717172	0.162983126905864\\
0.181818181818182	0.170172294279474\\
0.191919191919192	0.176660494771125\\
0.202020202020202	0.182370377483423\\
0.212121212121212	0.187243222099835\\
0.222222222222222	0.191246675955564\\
0.232323232323232	0.19438155632701\\
0.242424242424242	0.19668673641719\\
0.252525252525253	0.198241193576458\\
0.262626262626263	0.199162507164113\\
0.272727272727273	0.199601442248005\\
0.282828282828283	0.199732709188424\\
0.292929292929293	0.199742490723868\\
0.303030303030303	0.199813806067023\\
0.313131313131313	0.200111161933524\\
0.323232323232323	0.200766159679441\\
0.333333333333333	0.201865742973021\\
0.343434343434343	0.203444566587973\\
0.353535353535354	0.205482558806145\\
0.363636363636364	0.207908179949798\\
0.373737373737374	0.210607212033666\\
0.383838383838384	0.213436228196891\\
0.393939393939394	0.216239270584753\\
0.404040404040404	0.218865794981624\\
0.414141414141414	0.221187692569795\\
0.424242424242424	0.223113226701384\\
0.434343434343434	0.224596048723308\\
0.444444444444444	0.225638065793819\\
0.454545454545455	0.226285764208166\\
0.464646464646465	0.226620538184716\\
0.474747474747475	0.226744495467997\\
0.484848484848485	0.226763951488632\\
0.494949494949495	0.226773240273669\\
0.505050505050505	0.226841464234035\\
0.515151515151515	0.227004348860725\\
0.525252525252525	0.227262520223992\\
0.535353535353535	0.227586422747469\\
0.545454545454545	0.227926940172287\\
0.555555555555556	0.228229791275314\\
0.565656565656566	0.228451133904991\\
0.575757575757576	0.228571646958443\\
0.585858585858586	0.228606694111854\\
0.595959595959596	0.228610927913258\\
0.606060606060606	0.228676707875636\\
0.616161616161616	0.228926775734529\\
0.626262626262626	0.229502550463094\\
0.636363636363636	0.230550015197508\\
0.646464646464647	0.232205383370566\\
0.656565656565657	0.234582553537171\\
0.666666666666667	0.237763869433701\\
0.676767676767677	0.241795022409064\\
0.686868686868687	0.246684213539315\\
0.696969696969697	0.252405063486964\\
0.707070707070707	0.258902311479176\\
0.717171717171717	0.26609912228643\\
0.727272727272727	0.273904814676655\\
0.737373737373737	0.282221992369913\\
0.747474747474748	0.290952334936111\\
0.757575757575758	0.300000624669332\\
0.767676767676768	0.309276890565093\\
0.777777777777778	0.318696805144643\\
0.787878787878788	0.328180657152204\\
0.797979797979798	0.337651342759025\\
0.808080808080808	0.347031879544196\\
0.818181818181818	0.35624296386287\\
0.828282828282828	0.365201073210653\\
0.838383838383838	0.37381756499888\\
0.848484848484849	0.381999140103005\\
0.858585858585859	0.38964991887008\\
0.868686868686869	0.396675215603471\\
0.878787878787879	0.402986897815305\\
0.888888888888889	0.40850999177486\\
0.898989898989899	0.413189970536318\\
0.909090909090909	0.416999968602464\\
0.919191919191919	0.419947046649375\\
0.929292929292929	0.422076614435796\\
0.939393939393939	0.423474231667448\\
0.94949494949495	0.424264246664842\\
0.95959595959596	0.42460507877703\\
0.96969696969697	0.424681357500747\\
0.97979797979798	0.424693538106283\\
0.98989898989899	0.424845954180036\\
1	0.425334483932571\\
};
\addplot [color=black,mark size=1.5pt,only marks,mark=*,mark options={solid,fill=mycolor2,draw=mycolor2},forget plot]
  table[row sep=crcr]{%
0	0.356490294977174\\
};
\addplot [color=black,mark size=1.5pt,only marks,mark=*,mark options={solid,fill=mycolor2,draw=mycolor2},forget plot]
  table[row sep=crcr]{%
0.818456492386758	0.356490294977174\\
};
\addplot [color=black,mark size=1.5pt,only marks,mark=*,mark options={solid,fill=mycolor2,draw=mycolor2},forget plot]
  table[row sep=crcr]{%
0.818456492386758	0\\
};
\addplot [color=mycolor2,dashed,forget plot]
  table[row sep=crcr]{%
0	0.356490294977174\\
0.818456492386758	0.356490294977174\\
};
\addplot [color=mycolor2,dashed,forget plot]
  table[row sep=crcr]{%
0.818456492386758	0\\
0.818456492386758	0.356490294977174\\
};
\addplot [color=mycolor3,solid,forget plot]
  table[row sep=crcr]{%
0	0.425334483932571\\
1	0.425334483932571\\
};
\end{axis}
\end{tikzpicture}%